\documentclass[10pt,twocolumn,letterpaper]{article}
\usepackage[pagenumbers]{cvpr} 

\usepackage{graphicx}
\usepackage{amsmath}
\usepackage{amssymb}
\usepackage{booktabs}

\usepackage[pagebackref,breaklinks,colorlinks]{hyperref}

\usepackage[capitalize]{cleveref}
\crefname{section}{Sec.}{Secs.}
\Crefname{section}{Section}{Sections}
\Crefname{table}{Table}{Tables}
\crefname{table}{Tab.}{Tabs.}

\def\cvprPaperID{}
\def\confName{arXiv}
\def\confYear{2026}

\begin{document}

\title{Low-Cost Neural Radiance Fields}

\author{
Alice Huang\\
University of Illinois Urbana Champaign\\
{\tt\small aliceh4@illinois.edu}
\and
Prathamesh Sonawane\\
University of Illinois Urbana Champaign\\
{\tt\small pks10@illinois.edu}
\and
Yashdeep Thorat\\
University of Illinois Urbana Champaign\\
{\tt\small ythorat2@illinois.edu}
\and
Yug Rao\\
University of Illinois Urbana Champaign\\
{\tt\small yugrao2@illinois.edu}}

\maketitle

\begin{abstract}
Neural Radiance Fields (NeRF) achieve high-quality novel-view synthesis,
but their long training times and reliance on dense input views limit
accessibility. We present a comparative study of three accelerated NeRF
variants---DS-NeRF, TensoRF, and HashNeRF---and explore extensions targeted
at the low-compute, low-data regime. First, we add a depth-supervision loss
derived from COLMAP keypoints to TensoRF (TensoRF-DS) and evaluate it on
the LLFF dataset under reduced view counts. Second, we ablate the
feature-decoding MLP of TensoRF and study the effect of input downsampling
on PSNR and runtime on the synthetic Lego scene. Third, we propose four
architectural variants of the HashNeRF color and density networks,
including residual and convolutional designs, and report
PSNR/training-time tradeoffs under matched iteration budgets. Under
iso-time evaluation, none of our extensions conclusively outperform the
published baselines, but the experiments characterize which extensions
transfer to constrained settings and surface design questions for
future work.
\end{abstract}

\section{Introduction}
\label{sec:intro}

In this project, we implement and conduct experiments on the Neural Radiance Fields (NeRF) algorithm~\cite{mildenhall2020nerf}. This algorithm uses a Multilayer Perceptron Network whose input is a single continuous 5D coordinate, which contains information about spatial location and viewing direction, and whose output is the volume density and view-dependent emitted radiance at that spatial location(in terms of opacity and color). 

Formally, given inputs of a 3D coordinate $\boldsymbol{x}\in \mathbb{R}^3$ and viewing direction $\boldsymbol{d}\in \mathbb{R}^3$, NeRF outputs a density $\sigma$ and color $\boldsymbol{c}$ by deriving the function $f(\boldsymbol{x}, \boldsymbol{d}) = (\sigma, \boldsymbol{c})$. The output image is formed by casting rays $\boldsymbol{r} = \boldsymbol{o}+t\boldsymbol{d}$, with center of projection $\boldsymbol{o}$, direction $\boldsymbol{d}$, and parameter $t$. The color is modeled by:

$$\boldsymbol{\hat{C}}=\int_0^{\infty}T(t)\sigma(t)\boldsymbol{c}(t)dt,$$
$$T(t)=\exp({z-\int_0^t\sigma(s)ds})$$
Here, $T(t)$ is calculated to account for points earlier in the ray (i.e. from 0 to $t$) that would occlude the color at point $t$. The rendering loss of NeRF is the MSE loss over all of the rays for a given camera matrix $\mathbf{P}$.
$$\mathcal{L}_{\hat{C}} = \mathbb{E}_{\boldsymbol{r}\in\mathcal{R}(\mathbf{P})}||\boldsymbol{\hat{C}}(\boldsymbol{r})-\boldsymbol{C}(\boldsymbol{r})||_2^2$$

This model synthesizes 3D novel views of complex scenes given 2D images. Our focus is on NeRF models that improve performance and have reduced training and render time without sacrificing quality. We also look at how to decrease the training data needed to train a model.

A challenge that arises with NeRF is the fact that a model typically take over 10 hours to train with a large number of images. This can provide a challenge to those who do not have the time or computational resources to actually use NeRF. To address this, we explore existing NeRF variations that reduce the long training times and computational complexity that comes with NeRF and aim to make improvements to those variations as well. In the remainder of this paper, we will be discussing how we approach making these improvements (namely by using DS-NeRF \cite{kangle2021dsnerf}, TensoRF \cite{Chen2022ECCV}, and HashNeRF \cite{mueller2022instant}), our results, and finish with a discussion of our findings and future improvements.

\section{Related Work}

Here we will discuss DSNeRF, TensoRF, and HashNeRF, all extensions of NeRF. We explore these models by testing them and modifying them to see if additional improvements can be made.

\paragraph{DS-NeRF.}
One major limitation of NeRF is that it takes many images of the same scene in order to create a good radiance field reconstruction. Depth Supervised NeRF (DS-NeRF)~\cite{kangle2021dsnerf} attempts to improve the data efficiency of NeRF. The camera parameters of each scene are solved using the COLMAP algorithm~\cite{schoenberger2016mvs,schoenberger2016sfm}. These are used for calculating the rays for the model prediction that is then compared to the ground truth colors. Notably, in this process, COLMAP calculates 3D locations of keypoints. In DS-NeRF, this additional information is also used to train the model.

DS-NeRF adds a depth loss term in addition to the color loss. There are two different loss functions that are explored in the paper. One is derived from KL divergence, and the other is simple MSE loss. Both the KL loss and the MSE loss result in similar PSNR improvements, with the KL loss yielding slightly improved values.

While DS-NeRF improves PSNR with few images, it still takes a considerable number of iterations ($\mathcal{O}(10k-100k)$). In this project, we use this loss function training improvement to improve the low data PSNR of TensoRF~\cite{Chen2022ECCV}, which we cover next. Ideally, we can leverage the training time improvement of TensoRF with the low data performance of DS-NeRF in order to create a better model.

\paragraph{TensoRF.}
TensoRF \cite{Chen2022ECCV} offers a unique strategy for modeling and reconstructing radiance fields that sets it apart from NeRF, which relies solely on MLPs. Rather than solely focusing on MLPs, TensoRF treats the complete volume field as a 4D tensor and suggests decomposing it into numerous compressed, low-rank tensor components, resulting in a more efficient approach to scene modeling.

In our project, we explore the effect of downsampling on TensoRF to see if it speeds up computation and its effect on PSNR. We also perform an ablation study on the feature decoding architecture. We choose TensoRF since it is faster to train compared to NeRF and has lower space complexity as well – given our limited GPU and RAM resources, this model was an ideal pick.

\paragraph{HashNeRF.}
HashNeRF(Instant Neural Graphics Primitives with a Multiresolution Hash Encoding) \cite{mueller2022instant} proposes a radical approach to reducing training times for NeRF models using a hash encoding that maps inputs to a higher-dimensional space followed by a shallow neural network made up of fully connected layers. Moreover the multiresolution hash encoding enables the neural network to represent 3D geometry at multiple levels of detail. The paper proposes multiple hash tables for different resolutions and linear interpolation to embed features into the hash table.

In our project, we propose changes to the Neural network architectures that follows the hash encoding in order to get better results in the same number of iterations and similar training times. We have forked existing code for HashNeRF\cite{mueller2022instant}, and made modifications to the PyTorch model as well as added the functionality to generate runtime metrics and videos made using the novel views.

\section{Method}
\label{sec:approach}

\subsection{TensoRF-DS}
As discussed earlier, in our first approach, we add Depth Supervised training to TensoRF. To get the ground truth depth values, we extract the 3D point locations and 2D pixel coordinates of the COLMAP keypoints stored in the dataset. From this coordinate information and the camera matrix for the ground truth image, we can calculate the rays $\boldsymbol{r}(t) = \boldsymbol{o}+t\boldsymbol{d}$, along with a depth from the projection origin, $\boldsymbol{D}$. We then sample the model along these rays to calculate the model estimated depths given by 
$$\boldsymbol{\hat{D}}=\int_0^\infty T(t)\sigma(t)dt $$
We then use mean squared error to calculate the depth loss. While the DS-NeRF paper had better results with the KL loss, they were similar magnitude improvements, so we implement MSE loss as a baseline. We combine this depth loss with $\mathcal{L}_{\hat{C}}$, the NeRF color loss and weighting hyperparameter $\lambda_i$:
$$\mathcal{L}_{\hat{D}} = \mathbb{E}_{\boldsymbol{r}\in\mathcal{R}(\mathbf{P})}||\boldsymbol{\hat{D}}(\boldsymbol{r})-\boldsymbol{D}(\boldsymbol{r})||_2^2$$
$$\mathcal{L}_{\text{tot}} = \mathcal{L}_{\hat{C}}+\lambda_i\mathcal{L}_{\hat{D}}$$

We train this model on the LLFF dataset \cite{mildenhall2019llff}. This dataset consists of 20-30 images collected of household and office environments captured with a smartphone in a grid pattern. This dataset was thought to be the best fit for this approach because of small variance of angles, whereas other datasets have 360 views, which could be harder to train.

To carry out this approach, we fork the TensoRF repository~\cite{Chen2022ECCV} and add code to extract, predict, and compare ground truth and model depth outputs. We add the loss function calculation and also implement low data training. The specific details that yielded the results can be found in the next section.

\subsection{TensoRF}
We investigate the impact of downsampling on the performance of the TensoRF model in the context of computer vision tasks. Deep learning models such as TensoRF have shown remarkable performance in various computer vision applications such as image recognition, object detection, and semantic segmentation. However, these models require a vast amount of computational resources, including large amounts of memory and processing power.

We evaluate the effect of input downsampling on TensoRF's reconstruction quality, measured by train and test PSNR. Downsampling was also necessary to fit training within the memory budget of our hardware.

We also performed an ablation study on the feature decoding architecture of TensoRF. We did this by removing and adding layers and then training the model to see if those changes made any significant variation to the train and test PSNR. This ablation also serves as a probe of the underlying decoder architecture.

\subsection{HashNeRF}
The multi-level hash encoding model proposed by the paper \cite{mueller2022instant} has many applications like Gigapixel approximation, generating signed distance functions, Neural Radiance Caching. However, the model has been customized to be used for generating Neural Radiance fields. The architecture consists of 2 stages: Input encoding + Shallow Neural Network. Since the network is shallow, and has fewer parameters, the model trains faster and with the help of the hash encoding, the NeRF model also converges in fewer iterations as the encoding is also trainable alongside the neural network. Since the encoding is generic and trainable, it works with all kinds of Neural Graphics primitives - and hence can be used for varied tasks. The Neural Network model consists of two parts: Density MLP layers and Color MLP layers - the former outputs the log-space density, while the color layers add color variation information to the output.

Our work primarily focuses on experimenting with the Neural Network architectures, hyperparameters and trying to tweak the loss function the paper, that follows the Hash encoding. We tried various approaches to get better results and have analysed them using two metrics - PSNR (Peak Signal to noise ratio) which measures the ratio between the maximum possible value of a signal and the noise that affects the fidelity of its representation and the model's training time for the same number of iterations, since the lower time cost was the main highlight of this paper. Our main motivation for these experiments was to extensively try out different model architectures, loss function variations and types of layers to create better NeRF representations in roughly the same amount of training time. 

We attempted the following variations:
\begin{enumerate}
  \item Convolutional Layers: We attempted to replace the linear layers by 1D convolutional layers with a smaller kernel size in order to reduce the number of trainable parameters and make training faster.
  \item Residual connections from initial input to each layer in sigma and color MLP layers: We added a residual connection from the first layer by separating a part of the input and concatenating it to each following layer to preserve the spatial information from first layer.
  \item Adding residual connections from alternate layers: To improve upon the previous change, we added residual connections from 2 layers before each layer to preserve information from the previous layers while effectively tuning the parameters for each layer - which could not be done in the previous approach due to the residual from only first layer being added to each layer.
  \item Making the network deeper to fine tune the color variations: Added 3 linear layers along with residual from the sigma layers to further increase the number of parameters to get better color output.
\end{enumerate}

\section{Results}
\label{sec:results}

\subsection{TensoRF-DS}
The implementation of depth supervised TensoRF did not provide a significant improvement over the baseline. We first compare the overall PSNR results of the model. Three configurations were tested. First the full data was allowed to be trained. For an scene, there were 34 test images. 4 of these images served as test images, 29 were training images, and the last 1 image served as a validation image. The TensoRF model without depth supervision was used for the full data as a baseline. Then, 9 equally spaced images were selected, with 4 to train, 4 to test, and 1 for validation for the low data case. This case was tested both with and without depth supervision. The PSNR values for each of the configurations can be seen in \cref{tab:tensorfds}.  As seen in the table, there was a significant decrease in performance with the low data case, which was expected. However, there was only a very small increase of 0.01 in the PSNR with depth supervision.

\begin{figure}
  \centering
    \includegraphics[width=0.4\textwidth]{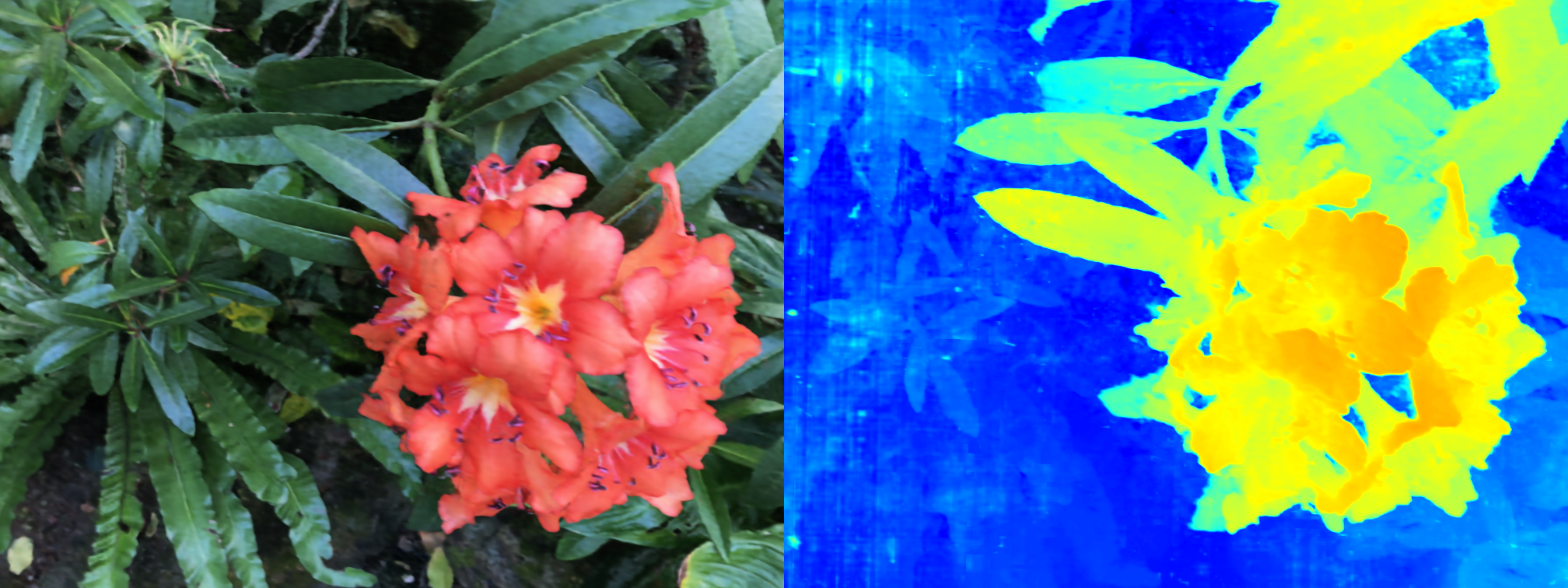}
    \includegraphics[width=0.4\textwidth]{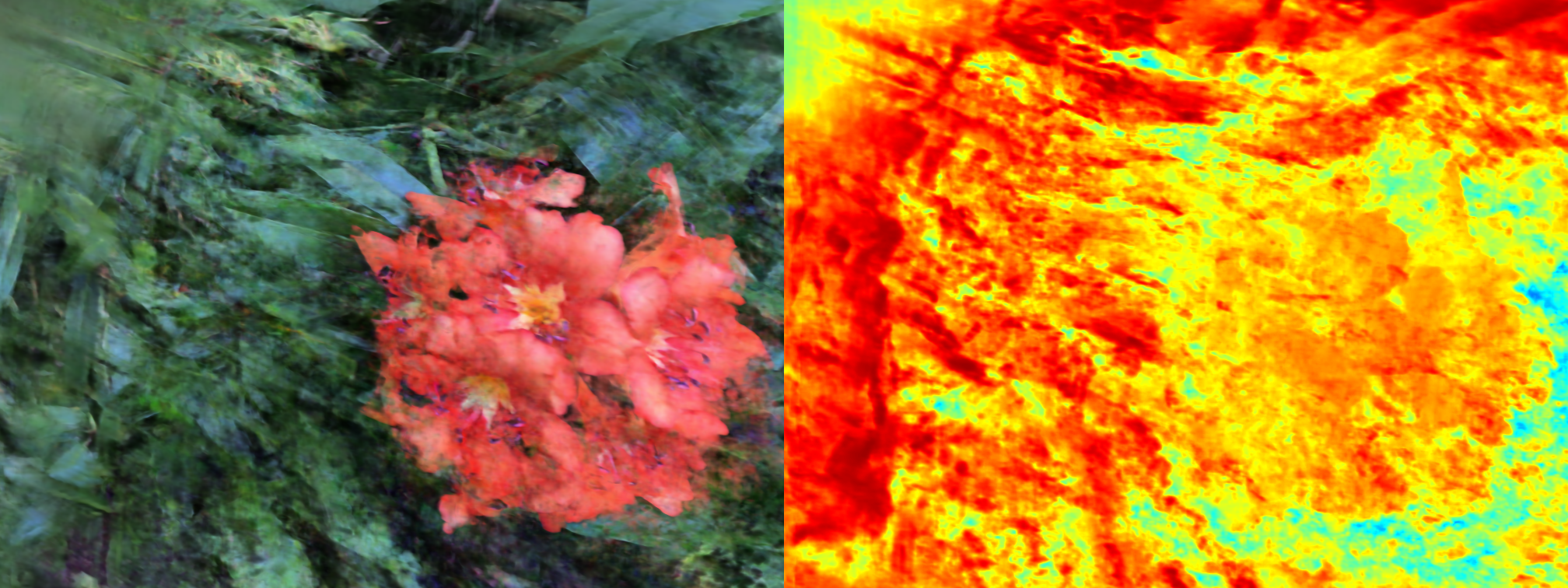}
    \includegraphics[width=0.4\textwidth]{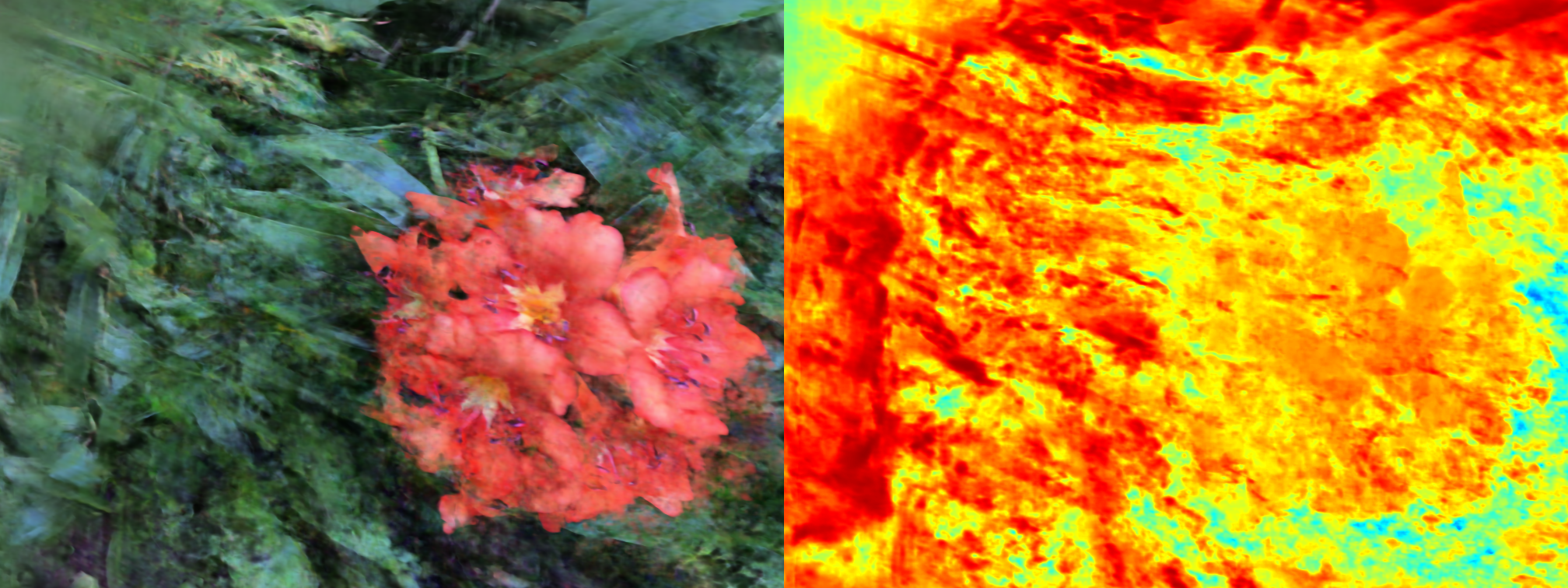}
  \caption{Output image (left) and depth map (right) for different configurations. Configurations from top to bottom: Full data TensoRF, Low data TensoRF, Low data TensoRF-DS \textit{Note: Though the bottom two images look similar, they are indeed different depth maps. We discuss a reason for this in \cref{sec:discconc}}}
  \label{fig:tdsimgs}
\end{figure}

These PSNR values serve as good quantitative measures for the success of the models, but we also have included visualizations that better show the differences between the models. These can be found in \cref{fig:tdsimgs}. in order to compare with the TensoRF baseline, all images were saved after 2500 iterations. In the figure, the color images are on the left and the depth images are on the right. The depth images show that while the color looks similar for the low data configuration, thee underlying depth structure is largely incomplete. Also, while both low data depth maps look very similar, they are indeed different as expected at the COLMAP keypoints, where the depth supervision was enforced. We discuss how this impacted the results and the mechanism of why this occurred in \cref{sec:discconc}.
\begin{table}
  \centering
  \begin{tabular}{@{}lcc@{}}
    \toprule
    Configuration & Train PSNR & Test PSNR \\
    \midrule
    Full Data TensoRF & 28.85 & 26.32 \\
    Low Data TensoRF & 29.92 & 15.23 \\
    Low Data TensoRF-DS & 29.93 & 15.23\\
    \bottomrule
  \end{tabular}
  \caption{Train and test PSNRs for all test configurations.}
  \label{tab:tensorfds}
\end{table}

\subsection{TensoRF}
We downsample both train and test data by a factor of 4 and train for 2{,}500 iterations (vs.\ the 30{,}000 used in the original TensoRF paper) due to memory and time constraints. We use the synthetic Lego scene from the NeRF dataset~\cite{mildenhall2020nerf} (100 training views, 200 test views).

Aligning with our expectations, downsampling the train and test datasets by a factor of 4 still resulted in good performance for our model. The model achieves the desired levels of accuracy, even given that we only performed 2{,}500 iterations instead of the default 30{,}000.

\begin{figure}[t]
  \centering
  \includegraphics[width=0.8\linewidth]{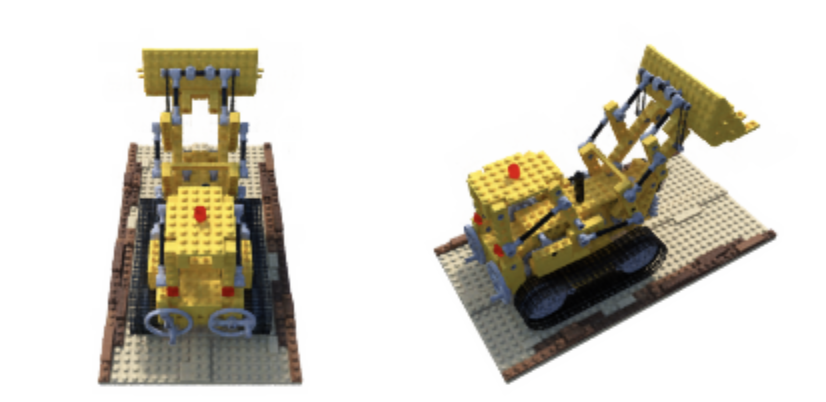}

   \caption{Output for TensoRF model using 2500 iterations and downsampling train and test by a factor of 4. These images achieve a train PSNR of 31.11 and test PSNR of 31.297.}
   \label{fig:tensorf-images}
\end{figure}

\cref{fig:tensorf-images} includes some visualizations of our model that trained for 2500 iterations on downsampled training and testing datasets (this model resulted in a train PSNR of 31.11 and a test PSNR of 31.297). This is using the default feature decoding architecture, which simply consists of 3 linear layers with ReLU activations in between. As you can see, the visualizations are quite clear and our PSNR values are sufficient – NeRF ~\cite{mildenhall2020nerf} achieves a PSNR of around 31 as well, but this model was also able to achieve that PSNR much faster and with similar performance (trained in less than 10 minutes).

We experimented with adding and removing linear layers to the feature decoding architecture but found that the changes we explored did not make significant improvements. \cref{tab:tensorf-table} contains an ablation table of architectures tried and corresponding PSNRs.

\begin{table}[t]
  \centering
\begin{center}
\begin{tabular}{@{}lcc@{}}
 \toprule
 Architecture Tried & Train PSNR & Test PSNR \\ [0.5ex]
 \midrule
 \midrule
 \vtop{
 \hbox{\strut Linear(in\_mlpC, featureC)}
 \hbox{\strut ReLU}
 \hbox{\strut Linear(featureC, featureC)}
 \hbox{\strut ReLU}
 \hbox{\strut Linear(featureC, 3)}}
 & 31.11 & 31.297 \\ 
 \midrule
 \vtop{
 \hbox{\strut Linear(in\_mlpC, featureC)}
 \hbox{\strut ReLU}
 \hbox{\strut Linear(featureC, featureC)}
 \hbox{\strut ReLU}
 \hbox{\strut Linear(featureC, featureC)}
 \hbox{\strut ReLU}
 \hbox{\strut Linear(featureC, 3)}}
 & 31.01 & 31.237 \\
 \midrule
 \vtop{
 \hbox{\strut Linear(in\_mlpC, featureC)}
 \hbox{\strut ReLU}
 \hbox{\strut Linear(featureC, 3)}}
  & 31.11 & 31.489 \\
 \midrule
 \vtop{
 \hbox{\strut Linear(in\_mlpC, featureC)}
 \hbox{\strut ReLU}
 \hbox{\strut Linear(featureC, 64)}
 \hbox{\strut ReLU}
 \hbox{\strut Linear(64, 32)}
 \hbox{\strut ReLU}
 \hbox{\strut Linear(32, 16)}
 \hbox{\strut ReLU}
 \hbox{\strut Linear(16, 8)}
 \hbox{\strut ReLU}
 \hbox{\strut Linear(8, 3)}} 
 & 30.65 & 30.923 \\
 \bottomrule
\end{tabular}
\end{center}
   \caption{Table of various feature decoding architectures tried and their corresponding train and test PSNR values. Note that these PSNR values were generated in 2500 iterations and by downsampling train and test by a factor of 4.}
   \label{tab:tensorf-table}
\end{table}

\subsection{HashNeRF}
All our models for HashNeRF have been trained on NVIDIA GeForce RTX 3090 Ti GPUs. We evaluated the model and results at 500 and 1000 iterations. Moreover, the dataset used for training, validation and testing consists of 100, 100 and 200 (RGB + depth) images respectively. 
From what we observed the default model architecture takes in the camera configuration details converts that into an image through one pipeline of and then further introduces color to these images through another pipeline of FC-layers. This architecture also consisted of residual connections to ensure that there is communication between the 2 pipelines and also to avoid vanishing gradients. 

Building on this intuition, we initially conducted experiments by introducing convolutional and dilated convolutional layers into the color generation pipeline. We hypothesized that providing the model with context of its surrounding pixels via convolutions could help it converge faster to the correct colors and improve accuracy. We tried various combinations of layers, accounting for over/under-fitting, vanishing gradients, and model flexibility. We also experimented with introducing convolutional layers into the image generation pipeline, but most of these variants underperformed the baseline reported by M\"uller et al.~\cite{mueller2022instant} at matched compute.

Another set of experiments we conducted were to try to take the model architecture from the paper and increasing the depth of each to the 2 pipelines to see how it affected the performance of the models. We also experimented with introducing a 3rd pipeline of FC-layers to re-establish the shape of the Nerf body once color was introduced the the picture. It followed a similar set of residual connections and layers as the other 2 pipelines but the motivation to this was as follows:
\begin{itemize}
    \item Pipeline 1: Take camera config as input and construct the basic shape/ outline of the object in the image (blue-print).
    \item Pipeline 2: Take the basic outline of image (blue-print) along with original camera config and guess the colors of the image.
    \item Pipeline 3 (our proposal): Take the basic outline of the image (blue-print), the colored Nerf representation and original camera configuration and make small corrections to the overall representation based on this information.
\end{itemize}

\begin{figure}[t]
  \centering
  \includegraphics[width=0.4\linewidth]{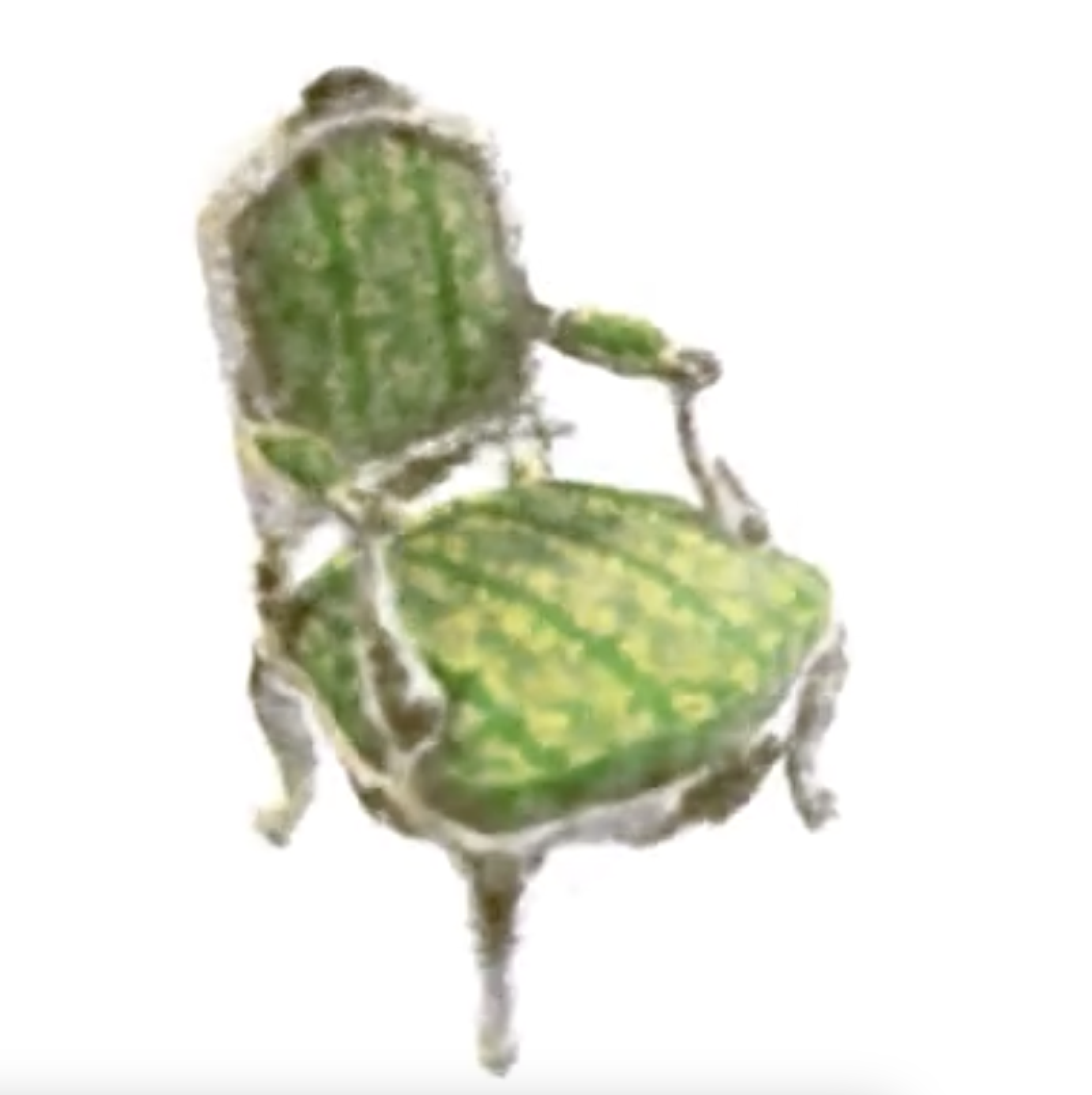}
  \includegraphics[width=0.4\linewidth]{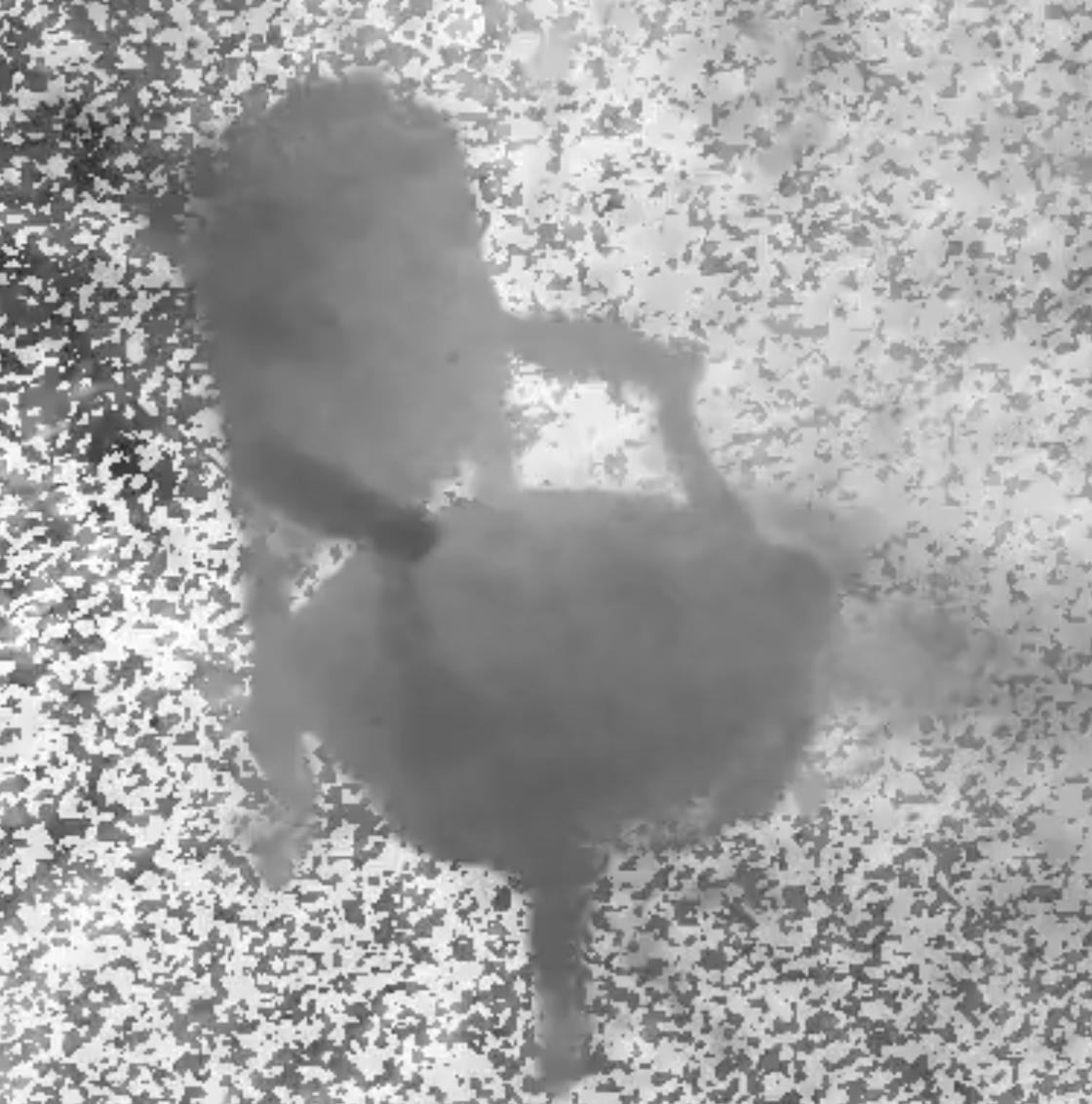}
   \caption{Output for HashNeRF model variation 4 using 1000 iterations and trained on the Synthetic Dataset. These novel views achieve a PSNR of 27.119.}
   \label{fig:hashnerf-images}
\end{figure}

\begin{table}[t]
  \centering
\begin{center}
\begin{tabular}{@{}lcc@{}}
 \toprule
 Architecture & PSNR & Training Time \\ [0.5ex]
 \midrule
 \midrule
 \vtop{
 \hbox{\strut Conv1d(48, 32, 3)}
 \hbox{\strut Conv1d(32, 16, 3)}
 \hbox{\strut Conv1d(16, 8, 3)}
 \hbox{\strut Conv1d(8, 4, 3)}}
 & 15.831 & 363.415 \\ 
 \midrule
 \vtop{
 \hbox{\strut Linear(32, 64, bias=False), ReLU}
 \hbox{\strut Concat(input views, output1)}
 \hbox{\strut Linear(79, 16, bias=False), ReLU}
 \hbox{\strut Concat(input views, output2)}
 \hbox{\strut Linear(31, 64, bias=False), ReLU}
 \hbox{\strut Concat(input views, output3)}
\hbox{\strut Linear(79, 64, bias=False), ReLU}
\hbox{\strut Concat(input views, output4)}
\hbox{\strut Linear(79, 3, bias=False)}
}
 & 25.873 & 306.751 \\
 \midrule
 \vtop{
 \hbox{\strut Linear(32, 64, bias=False), ReLU}
 \hbox{\strut Linear(79, 16, bias=False), ReLU}
 \hbox{\strut Concat(input views, output1)}
 \hbox{\strut Linear(31, 64, bias=False), ReLU}
 \hbox{\strut Concat(output1, output2)}
\hbox{\strut Linear(79, 64, bias=False), ReLU}
\hbox{\strut Concat(output2, output3)}
\hbox{\strut Linear(79, 3, bias=False)}
}
  & 26.547 & 305.208 \\
 \midrule
 \vtop{
 \hbox{\strut Linear(32, 64, bias=False), ReLU}
 \hbox{\strut Linear(64, 16, bias=False)}
 \hbox{\strut Concat(input views, output1)}
 \hbox{\strut Linear(20, 64, bias=False), ReLU}
\hbox{\strut Linear(64, 64, bias=False)}
\hbox{\strut Concat(input views, output2)}
\hbox{\strut Linear(50, 128, bias=False), ReLU}
\hbox{\strut Linear(128, 64, bias=False), ReLU}
\hbox{\strut Linear(64, 3, bias=False)}}
 & 27.119 & 319.669 \\
 \bottomrule
\end{tabular}
\end{center}
   \caption{Table of various shallow architectures for HashNeRF and their corresponding PSNR values and training times. Note that these PSNR values were generated in 1000 iterations.}
   \label{tab:hashnerf-table1}
\end{table}

\section{Discussion and Future Work}
\label{sec:discconc}

\subsection{TensoRF-DS}
These experiments showed that there was not very good improvement with the Depth Supervised TensoRF approach. There is low PSNR with low data as expected, but there is very minimal (~0.01 PSNR) increase between the non-depth supervised and the depth supervised approaches. We now discuss possible reasons for this from some analysis and literature reading.

Initially, it was thought that the depth supervised result and the regular low data training approach were yielding the same depth maps. Upon further investigation, it was found that there were indeed some improvements being made. However, there were orders of magnitude fewer COLMAP key points than there were color pixel comparisons in the other loss term. Because of this, the rays for the depth supervision were only intersecting a small portion of the voxel space. This means that while the depth loss term gradients were being correctly propagated to change the $\sigma$ values, only a small number of the components were being affected. To solve this, in a future project, it would be possible to calculate the 3d ground truth mesh at more points and use a ground truth scene depth map similar to the color map used.

A further limitation is that our experiments used a constrained training budget; while this did not hinder the baseline models, it is plausible that the depth-supervised model was simply not trained for long enough. DS-NeRF resulted in a 2-3x improvement over NeRF, but it still used $\mathcal{O}(10k)$ iterations. While the change described earlier may solve the problem with the model, increasing the iterations may also fix the problem with the current model.

These possible future approaches could be used to train a model that successfully leverages the training time and memory improvements of TensoRF as well as the low data reconstruction quality of DS-NERF.

\subsection{TensoRF}
The experimental results demonstrate that downsampling can be an effective technique for improving computational efficiency without significantly compromising model quality. Specifically, our findings suggest that downsampling the training and test datasets by a factor of 4 yields comparable peak signal-to-noise ratio (PSNR) values to those achieved by the full model.

These results are encouraging for researchers who seek to employ TensoRF in their work but may lack the necessary computational resources to train the model at full scale. By applying appropriate levels of downsampling, researchers can still achieve satisfactory results while reducing the computational burden.

Moreover, our investigation of different feature decoding architectures did not result in significant improvements, which suggests that the current architecture is already capable of generating high-quality results. 

In the future, a more thorough ablation study on different parts of the TensoRF architecture could yield improvements and insights into the model. Future work could also extend the iteration budget to verify whether the gap to full-resolution training closes.

Overall, these results highlight the potential of downsampling as a practical and effective technique for scaling up TensoRF and facilitating its use in various applications.

\subsection{HashNeRF}
This architecture outperformed our other variants (\cref{tab:hashnerf-table1}). At a matched 1{,}000-iteration budget, it achieves a PSNR of 27.12, compared to the 28.2 reported by M\"uller et al.~\cite{mueller2022instant} under their default configuration. Our network has slightly more parameters than the original; when allowed to train to convergence (rather than to a fixed time budget), it surpassed the original PSNR, but we report only the iso-time comparison here as the fair baseline. We additionally experimented with replacing the L2 photometric loss with mean absolute error and Huber loss, but observed no improvement in any case.

We also conducted experiments replacing the FC layers with sequence-to-sequence modules to study their effect on the model. We were unable to test beyond 5--6 architectures for this part due to resource and time constraints introduced by the higher parameter counts. The results from this limited exploration did not match the PSNR reported by M\"uller et al.~\cite{mueller2022instant}.

\section{Conclusion}
\label{sec:conclusion}
We presented a comparative study of three accelerated NeRF variants under constrained training budgets, contributing a depth-supervised extension of TensoRF, an ablation of its feature-decoding MLP, and four architectural variants of the HashNeRF color and density networks. Under iso-time evaluation, none of our modifications conclusively outperformed the published baselines; however, the experiments characterize which extensions transfer to limited-compute settings and motivate the design questions raised above. Closing the gap to the original baselines under matched compute remains an open direction for future work.

{\small
\bibliographystyle{ieee_fullname}
\bibliography{egbib}
}

\end{document}